\documentclass[sigconf]{acmart}
\usepackage[capitalize]{cleveref}
\usepackage{graphicx}
\usepackage{amsmath}
\usepackage{booktabs}
\usepackage{pifont}
\usepackage{amsmath}
\usepackage{booktabs}
\usepackage{multirow}
\usepackage{bbding}
\usepackage{tikz}
\usepackage{colortbl}
\usepackage{balance}
\Crefname{algocf}{Algorithm}{Algorithms}
\usepackage[ruled,linesnumbered]{algorithm2e}
\usepackage{algpseudocode}

\definecolor{Gray}{gray}{0.9}
\definecolor{LightCyan}{rgb}{0.88,1,1}

\newcolumntype{a}{>{\columncolor{Gray}}c}
\newcolumntype{b}{>{\columncolor{white}}c}

\AtBeginDocument{%
  }

\acmConference[MM '23] {Proceedings of the 31st ACM International Conference on Multimedia}{October 29--November 3, 2023}{Ottawa, ON, Canada.}
\copyrightyear{2023}
\acmYear{2023}
\setcopyright{acmlicensed}\acmConference[MM '23]{Proceedings of the 31st ACM International Conference on Multimedia}{October 29-November 3, 2023}{Ottawa, ON, Canada}
\acmBooktitle{Proceedings of the 31st ACM International Conference on Multimedia (MM '23), October 29-November 3, 2023, Ottawa, ON, Canada}
\acmPrice{15.00}
\acmDOI{10.1145/3581783.3612297}
\acmISBN{979-8-4007-0108-5/23/10}

\acmSubmissionID{2662}

\begin{document}

\title{ProtoHPE: Prototype-guided High-frequency Patch Enhancement \\ for
Visible-Infrared Person Re-identification}

\author{Guiwei Zhang}
\affiliation{%
  \institution{Beijing Key Laboratory of Digital Media, School of Computer Science and Engineering, Beihang Univerisity.}
  \city{Beijing}
  \country{China}}
\email{zhangguiwei@buaa.edu.cn}

\author{Yongfei Zhang}
\authornote{Corresponding author.}
\affiliation{%
  \institution{Beijing Key Laboratory of Digital Media, State Key Laboratory of Virtual Reality Technology and Systems, School of Computer Science and Engineering, Beihang Univerisity.}
  \city{Beijing}
  \country{China}}
\email{yfzhang@buaa.edu.cn}

\author{Zichang Tan}
\affiliation{%
  \institution{Department of Computer Vision Technology (VIS), Baidu Inc.}
  \city{Beijing}
  \country{China}}
\email{tanzichang@baidu.com}

\renewcommand{\shortauthors}{Guiwei Zhang, Yongfei Zhang, \& Zichang Tan}

\begin{abstract}
Visible-Infrared person re-identification is challenging due to  the large modality gap. To bridge the gap, most studies heavily rely  on the correlation of visible-infrared holistic person images, which may perform poorly under severe distribution shifts.  In contrast, we find that some cross-modal correlated high-frequency components  contain discriminative visual patterns and are less affected by variations such as wavelength, pose, and background clutter than holistic images. Therefore, we are  motivated to  bridge the modality gap based on such high-frequency components, and propose  \textbf{Proto}type-guided \textbf{H}igh-frequency \textbf{P}atch \textbf{E}nhancement (ProtoHPE) with two core designs. \textbf{First}, to enhance the representation ability  of cross-modal correlated high-frequency components, we split patches with such  components by Wavelet Transform and exponential moving average Vision Transformer (ViT), then
 empower ViT to take the split  patches as auxiliary input.
 \textbf{Second}, to obtain semantically compact and discriminative high-frequency representations of the same identity, we propose Multimodal Prototypical Contrast. To be specific, it hierarchically captures  comprehensive semantics of different modal instances,
 facilitating the aggregation of  high-frequency representations belonging to the same identity.  With it,
 ViT can  capture key high-frequency components during inference without relying on ProtoHPE, thus bringing no extra complexity. Extensive experiments  validate the effectiveness of ProtoHPE.
\end{abstract}




\received{20 February 2007}
\received[revised]{12 March 2009}
\received[accepted]{5 June 2009}

\maketitle

\section{Introduction}
\begin{figure}[h]
  \centering
  \includegraphics[width=0.95\linewidth]{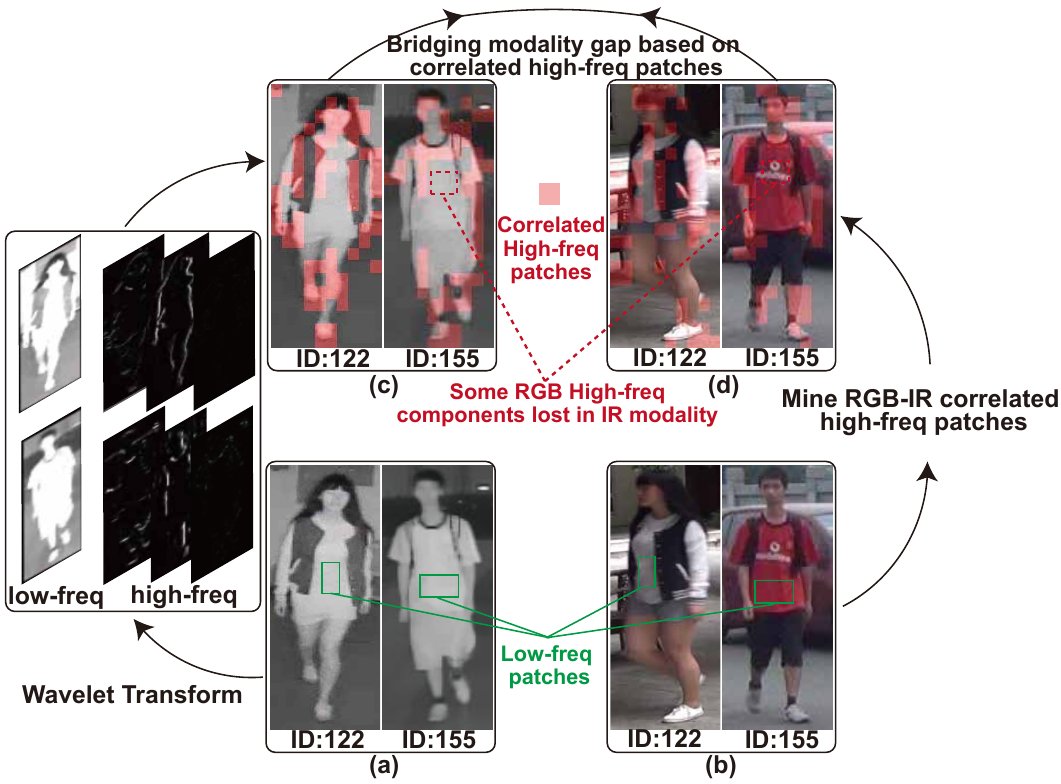}
  \caption{Illustration of our idea. (a) and (b) are holistic IR and RGB  images, respectively. In (c) and (d), some RGB-IR correlated high-frequency patches contain discriminative visual patterns, e.g., heads, cloth textures, and human silhouettes. Such patches are less affected by variations such as wavelength, pose, and background clutter than holistic images, and thereby are more
  robust to  distribution shifts.
  We are thus  motivated to
  bridge the gap based on these patches.}
\label{intro}
\end{figure}
Person re-identification (ReID) aims to retrieve a specific person, given in a query image, from non-overlapping camera views~\cite{ye2021deep,zhang2021unrealperson,zhang2020single}. Most research to date has focused on RGB cameras~\cite{he2021transreid,zhu2022part,tan2022dynamic,zhang2023pha}. However, RGB  cameras perform poorly in capturing  the appearance  of people in low-light conditions. In contrast, near-infrared (IR) cameras can remedy such inadequacy of RGB cameras due to their robustness to illumination variations. Hence, visible-infrared ReID (VI-ReID) has recently attracted great attention. Specifically, VI-ReID aims to retrieve a  person, given in an IR query, from the search over RGB person images and vice versa~\cite{ye2021deep,wang2020cross}.

VI-ReID is challenging due to the large modality gap. To bridge the  gap, most  existing methods focus on learning modality-shared and modality-compensated
 feature representations~\cite{zhang2022modality,lu2022learning,jiang2022cross}. However, these methods heavily rely on the correlation
 of RGB and IR holistic person images  and may perform poorly under severe distribution shifts. Such distribution shifts are common in VI-ReID due to large differences in wavelength, scattering, pose, and background clutter between person images of different modalities. This issue was still  understudied in previous VI-ReID works, thus resulting in  poor performance in bridging the modality gap.

 To this end, we explore a more suitable solution to bridge the modality gap  from a frequency perspective, which is of great significance in digital image processing~\cite{fujieda2018wavelet,yao2022wave,miao2023f,miao2022hierarchical,zhang2023pha}. In Fig.1, we first utilize  Haar Wavelet Transform to decompose IR  person images into low- and high-freq components, and then select patches with top-$K$ high-freq responses, termed IR high-freq patches.\footnote[1]{\textbf{For simplicity, we abbreviate ``low-frequency'' and ``high-frequency'' as ``low-freq'' and ``high-freq'',  respectively, in this paper.}} Please note that we directly define the high-freq responses as the $l_2$-norm of  high-freq components of person images (\cref{CHPE}).
 We compare the selected patches above and reveal:

 \textbf{(1) IR  high-freq patches contain more discriminative visual patterns than IR low-freq patches.} From ~\cref{intro} (a), due to longer wavelength and more scattering than visible light, visually similar IR low-freq patches can represent different  semantics, e.g., visual patterns of  white and red shirts in ~\cref{intro} (b).
 Enhancing such patches may cause feature representations to lose identity discrimination. In contrast,
 IR high-freq patches contain more discriminative visual patterns, e.g., heads, certain cloth textures, and human silhouettes in ~\cref{intro} (c), and thus are more critical components to boost  VI-ReID.

 \textbf{(2) The modality gap between RGB-IR correlated high-freq patches is much smaller than that between holistic images.} From  ~\cref{intro} (c) and (d), although some RGB high-freq components are lost in the IR modality due to large differences in wavelength and scattering between two modalities,
 there are always RGB patches  highly correlated with IR high-freq patches. Such  RGB-IR correlated high-freq patches  contain discriminative visual patterns that are less affected by variations such as wavelengths, pose, and background clutter than holistic images, and thereby are more robust to the distribution shift. We are thus motivated to bridge the modality gap based on RGB-IR correlated  high-freq patches effectively.

However, previous VI-ReID methods~\cite{lu2020cross, wei2021syncretic,lu2022learning} only bridge the modality gap by optimizing instance representations in the mini-batch, which may cause inconsistent representations of the same identity in different mini-batches. To address this issue,  recent research  proposes cross-modal instance-prototype contrast ~\cite{chen2022prototypical}, in which a prototype is defined as a representative embedding for a group of semantically similar instances. Despite progress, this contrastive approach ignores direct semantic constraints between different modal prototypes. Since prototypes can
characterize the joint distribution of multiple modalities in a compact form,  we argue that enforcing direct semantic constraints between prototypes facilitates stable interactions  between structural distributions of different modalities. This is beneficial  to extract more compact and informative  representations, which improves robustness to the gap.

In summary, the key to our idea lies in two points: capturing RGB-IR correlated  high-freq patches, and robustly bridging the modality gap based on  these patches. To this end, we propose \textbf{Proto}type-guided \textbf{H}igh-freq \textbf{P}atch \textbf{E}nhancement (ProtoHPE) with two core designs. \textbf{First},
 we propose  Cross-modal High-freq Patch Enhancement (\cref{CHPE}), in which Haar Wavelet Transform and exponential moving average Vision Transformer (ViT)~\cite{dosovitskiy2020image} are jointly utilized to mine RGB-IR correlated high-freq patches. Such patches are then fed into ViT as
 auxiliary input to
 enhance the representation ability of  RGB-IR correlated high-freq components. \textbf{Second}, to obtain semantically compact and discriminative  high-freq representations of the same identity  during network optimization, we propose Multimodal Prototypical Contrast (MultiProCo).
 % It upgrades traditional instance-level discrimination to multimodal prototype-level discrimination.
 Note that a modality prototype is defined as a representative embedding for instances belonging to  the same identity and the same modality.
 MultiProCo (\cref{multiproco-section}) captures comprehensive semantics of different modal instances in a hierarchical manner, where lower-level {\it instance-prototype constraint} captures the semantics of  instances within a modality belonging to the same identity,  while higher-level {\it multimodal prototypes contrastive regularization} captures structural distributions across different modalities and different identities.  This facilitates the  aggregation of  high-freq enhanced representations belonging to  the same identity, which helps to learn compact and discriminative representations to robustly bridge the gap.
 % This benefits the ViT  to
 % extract robust and discriminative high-frequency representations of the same identity, thereby bridging the modality gap stably.
 % By virtue of MultiProCo, our ProtoHPE can still capture critical  high-frequency components even with  the entire sequence as input.
 With  MultiProCo, ViT can  capture key high-freq components during inference without relying on ProtoHPE, thus bringing no extra complexity.
Our contributions include:
\begin{itemize}
\item[$\bullet$] We propose Cross-modal High-freq Patch Enhancement to enhance the representation ability  of RGB-IR correlated high-freq patches. Bridging the modality gap based on these patches is more robust to the distribution shift compared to holistic images.

\item[$\bullet$] We propose  Multimodal Prototypical Contrast, which hierarchically captures  comprehensive semantics of different modal instances. It facilitates the aggregation of   high-freq enhanced representations belonging to the same identity,
thus bridging the  modality gap robustly. With it, ViT can  capture key high-freq components during inference without relying on ProtoHPE,     bringing no extra complexity.
  \item[$\bullet$] Extensive experimental results perform favorably against mainstream methods on SYSU-MM01 and RegDB datasets.
\end{itemize}

\section{Related Work}
\subsection{Visible-Infrared Person Re-identification}
Visible-Infrared person Re-identification is challenging due to the large cross-modal gap. To bridge the gap, existing schemes can be roughly summarized into the following two aspects.

(1) Some works mitigate the large modality discrepancy by introducing generation-based methods~\cite{wang2019rgb,wang2019learning,choi2020hi,li2020infrared}. AlignGAN~\cite{wang2019rgb} exploits pixel- and feature-level alignment jointly to alleviate the cross-modal  variations. $\textup{D}^2$RL ~\cite{wang2019learning} mitigates the modality discrepancy by generating multi-spectral images with a bi-directional cycle GAN. XIV~\cite{li2020infrared} generates an auxiliary X-modality to bridge the visible and infrared modalities.

(2) Others develop various dual-stream networks to  learn modal-shared and modal-compensated features~\cite{ye2020cross,lu2020cross,gao2021mso,zhang2021global,ye2020dynamic,zhang2022modality,fu2021cm,lu2022learning}. MACE~\cite{ye2020cross} proposes a sharable two-stream network to alleviate the modality gap in both feature and classifier levels. MSO~\cite{gao2021mso} proposes  edge features enhancement  to enhance  modality-sharable features. CM-NAS~\cite{fu2021cm} proposes BN-oriented search architecture to boost cross-modal matching. cm-SSFT~\cite{lu2020cross} explores the potential of   both modality-shared and modality-specific features  to boost performance. MSCLNet~\cite{zhang2022modality}  synergizes and complements instances of different modalities to learn discriminative representations. CMT~\cite{jiang2022cross} introduces a Transformer encoder-decoder design to compensate for the missing modality-specific information.

Despite inspiring progress, the above approaches heavily rely on the correlation of RGB-IR holistic person images and may perform poorly under severe distribution shifts. In contrast, we  bridge the modality gap based on RGB-IR correlated high-freq  patches, which  are more robust to distribution shifts than holistic images.

\subsection{Self-supervised Learning}
Self-supervised learning (SSL) for vision aims to learn general-purpose  representations without human supervision~\cite{ericsson2022self}. Since SimCLR ~\cite{chen2020simple} demonstrated  effectiveness on the instance-level discrimination task, contrastive learning has dominated the state-of-the-art  SSL models~\cite{chen2021empirical,caron2021emerging,li2020prototypical}.
% SimCLR maximizes the consistency of representations between different augmented views of the same image while minimizing the consistency of representations from different images.
However, SimLCR heavily relies  on sufficient negatives in the mini-batch. To tackle this issue, MoCo ~\cite{he2020momentum} introduces a dynamic dictionary with a queue and SimSiam~\cite{chen2021exploring} learns meaningful representations without negative sample pairs.

Recently, the success of contrastive learning  has inspired the generalization of this method to Contrastive Language Image Pretraining (CLIP)~\cite{radford2021learning}, where images and texts are considered as multimodal views of the same underlying concepts. X-CLIP~\cite{ma2022x} focuses on   multi-grained contrastive learning for accurate cross-modal alignment.
ProtoCLIP~\cite{chen2022prototypical} sets up instance-prototype discrimination, which efficiently transfers higher-level structural knowledge in vision-language pretraining.
Progressive self-distillation~\cite{andonian2022robust}  learns robust representations from noisy data to boost cross-modal contrastive learning.

However, the above approaches neglect direct semantic constraints between different modal prototypes. Since prototypes can characterize the joint distribution of multiple modalities in a compact form, imposing direct semantic constraints between prototypes helps to capture structural distributions of different modalities. Therefore, we propose multimodal prototypical contrast,
which hierarchically captures  comprehensive semantics of different modal instances. This facilitates learning compact and discriminative representations, which improves  robustness to the modality gap.

\section{Preliminaries}
\label{baseline}
\subsection{Part-based ViT Baseline.}
Given a RGB/IR  person image,
 we split it into $N$ patches $\{\boldsymbol{x}_i|i=1,2,\cdots,N\}$ and tokenize  them  with learnable linear projection $\mathcal{F}$ and position embeddings $\mathbf{E}_{\mathrm{pos}}\in {\mathbb{R}}^{(1+N)\times D}$, such that
 \begin{equation}
 \setlength{\abovedisplayskip}{4pt}
\setlength{\belowdisplayskip}{4pt}
 \small
\mathbf{X}=\left[\boldsymbol{x}_{[\mathrm{CLS}]}, \mathcal{F}({\boldsymbol{x}_1}), \mathcal{F}({\boldsymbol{x}_2}), \cdots, \mathcal{F}(\boldsymbol{x}_N)\right]+\mathbf{E}_{\mathrm{pos}}
\label{one}
\end{equation}
where $\mathbf{X} \in {\mathbb{R}}^{(1+N)\times D}$ denotes the input sequence and  $\boldsymbol{x}_{[\mathrm{CLS}]}$ is a learnable class token. We take  $\mathbf{X}$ as input to a sequence of $l$ transformer blocks, and the hidden output of the $l$-th block is denoted by $\boldsymbol{h}_{l}$.
Subsequently, all embeddings in $\boldsymbol{h}_{l}$ except the class token
$\boldsymbol{x}_{[\mathrm{CLS}]}$ are divided into $T$ parts uniformly. Then, $T$ parts that concatenate $\boldsymbol{x}_{[\mathrm{CLS}]}$ are further fed into a shared Transformer block to extract $T$ local part representations $\left\{\boldsymbol{f}_{j} \mid j=1,2, \cdots, T\right\}$, in which $\boldsymbol{f}_{j}$ is the encoded class token of the $j$-th part. Furthermore, a global Transformer block operates on the hidden output $\boldsymbol{h}_{l}$, and the encoded class token is served as the global representation $\boldsymbol{f}_{g}$.
By convention, we optimize the representations $\boldsymbol{f}_{g}$ and  $\left\{\boldsymbol{f}_{j} \mid j=1,2, \cdots, T\right\}$ with the
cross-entropy loss $\mathcal{L}_{CE}$ and the triplet loss $\mathcal{L}_{Tri}$, respectively:
\begin{equation}
\small
\setlength{\abovedisplayskip}{4pt}
\setlength{\belowdisplayskip}{4pt}
\begin{split} \mathcal{L}_{base}=\mathcal{L}_{CE}\left(\boldsymbol{f}_{g}\right)+\mathcal{L}_{Tri}\left(\boldsymbol{f}_{g}\right)+
    \frac{1}{T} \sum_{j=1}^T\left(\mathcal{L}_{CE}\left({\boldsymbol{f}_{j}}\right)+\mathcal{L}_{Tri}\left(\boldsymbol{f}_{j}\right)\right)
\label{con:baseline}
\end{split}
\end{equation}
$\mathcal{L}_{base}$ benefits the ViT to elevate the discriminative power of both  global and fine-grained part representations.
\subsection{Prototypical Contrastive Learning.} The objective is to embed the positive pair nearby in the representation space while embedding negative pairs far apart. Note that each positive pair consists of an instance and its associated semantic prototype, while  negative pairs consist of paired instances with unrelated semantic prototypes. Specifically, given a positive pair $(v,c)$, the ProtoNCE loss~\cite{li2020prototypical} is formulated below:
\begin{equation}
\small
\mathcal{L}_{\operatorname{ProtoNCE}}\left(v, c, \mathcal{N}_c,\left\{\tau_c\right\}\right)=-\log \frac{\exp \left(v \cdot c / \tau_c\right)}{\sum_{c_j \in\{c\} \cup \mathcal{N}_c} \exp \left(v \cdot c_j / \tau_{c_j}\right)}
\end{equation}
where $\mathcal{N}_c$ denotes a set of negative prototypes for instance representation $v$, and $\tau_c$ is a prototype-specific temperature parameter.

\section{Method}
\begin{figure*}[h]
  \centering
  \includegraphics[width=0.9\textwidth]{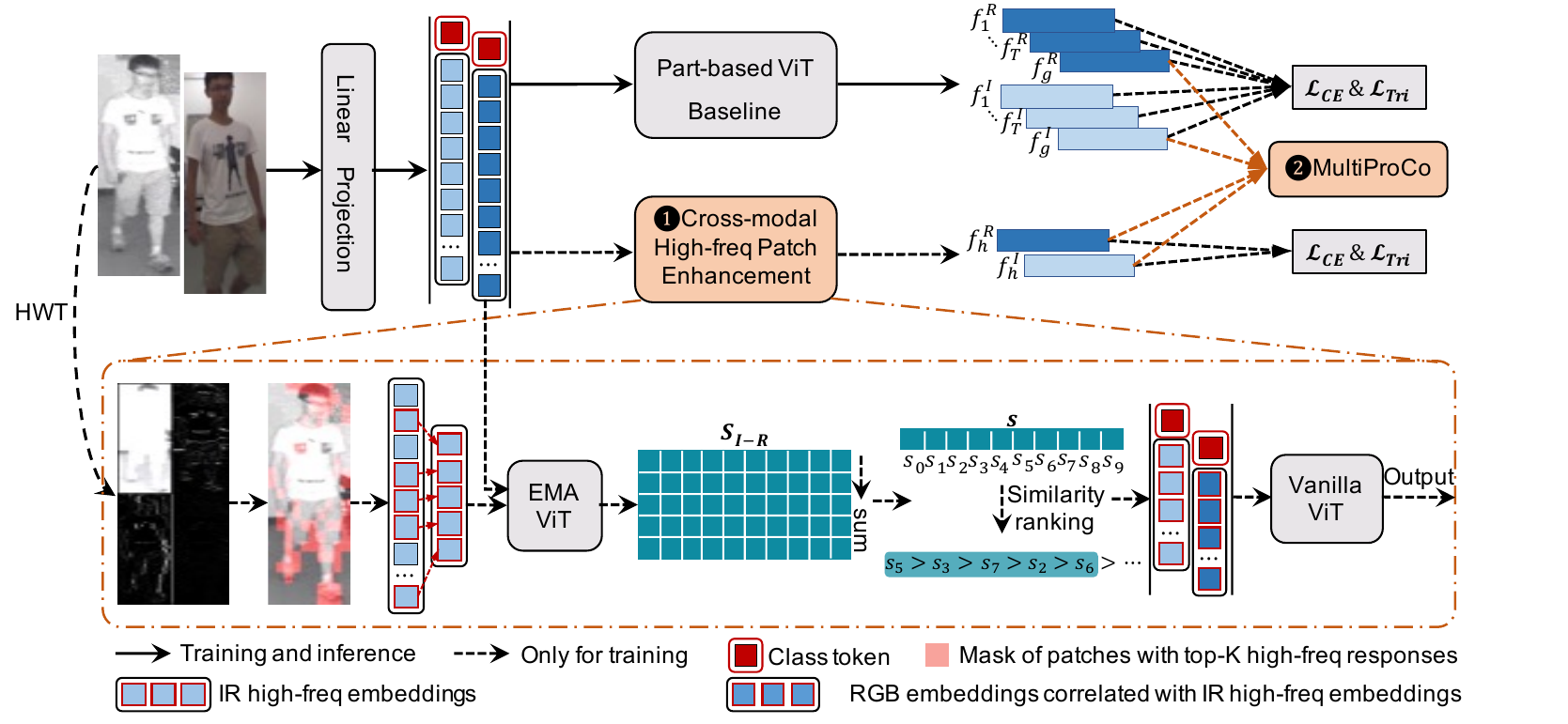}
  \caption{The overall of  ProtoHPE, including  \ding{202} Cross-modal High-freq Patch Enhancement (CHPE) and \ding{203} Multimodal Prototypical Contrast (MultiProCo).  In CHPE, Haar Wavelet Transform (HWT) and an Exponential Moving Average ViT (EMA ViT) are jointly utilized to mine RGB-IR correlated high-freq patches. Such patches are then fed into vanilla ViT to
 enhance the representation ability of  RGB-IR correlated high-freq components. Note that vanilla ViT and Part-based ViT share weights during training.
 MultiProCo hierarchically captures  comprehensive semantics of different modal instances, facilitating the  aggregation of  global and high-freq enhanced representations belonging to  the same identity. With it,
 The baseline can  capture key high-freq components during inference without relying on ProtoHPE, thus bringing no extra complexity.
  }
\label{overview}
\end{figure*}

\cref{overview} shows the overall pipeline of  Prototype-guided High-frequency Patch Enhancement (ProtoHPE), including \ding{202} Cross-modal High-freq Patch Enhancement (CHPE)  and \ding{203} Multimodal Prototypical Contrast (MultiProCo). CHPE (\cref{CHPE}) aims to enhance the representation ability of RGB-IR correlated high-freq patches. Then, both global and high-freq enhanced representations are regularized by MultiProCo (\cref{multiproco-section}). MultiProCo hierarchically captures comprehensive semantics of different modal instances, facilitating the aggregation of representations belonging to the same identity.
\subsection{Cross-modal High-freq Patch Enhancement}
\label{CHPE}
The objective  is to enhance the representation ability of RGB-IR correlated high-freq patches that  are more robust to the distribution shift than holistic images. Below, we explain the underlying mechanism of CHPE, which consists of \textbf{(\romannumeral1) IR high-freq sampling} and  \textbf{(\romannumeral2) RGB-IR correlated high-freq patch enhancement}.

 \textbf{(\romannumeral1) IR  high-freq sampling.}
Given an IR person image $\boldsymbol{I} \in \mathbb{R}^{\it H\times W\times C}$, where {\it H, W} and {\it C} represent its height, width and the number of channels respectively,  we  utilize  Haar Wavelet Transform to decompose it into four wavelet subbands: ${\boldsymbol{I}_{LL}}$, ${\boldsymbol{I}_{LH}}$, ${\boldsymbol{I}_{HL}}$, and  ${\boldsymbol{I}_{HH}} \in \mathbb{R}^{H/2 \times W/2 \times C}$. Note that ${\boldsymbol{I}_{LL}}$ mainly retains low-freq components that depict the overall appearance of a person image, while ${\boldsymbol{I}_{LH}}$, ${\boldsymbol{I}_{HL}}$, and  ${\boldsymbol{I}_{HH}}$ reflect high-freq components that contain discriminative visual patterns such as  heads and human silhouettes. Thus, we sum the later three subbands together:
\begin{gather}
\small
\mathcal{M}_h({\boldsymbol{I}})=\operatorname{sum}\left(\boldsymbol{I}_{LH}, \boldsymbol{I}_{HL}, \boldsymbol{I}_{HH}\right), \mathcal{M}_h({\boldsymbol{I}})\in\mathbb{R}^{H/2\times W/2\times C}
\end{gather}
We further introduce a projection function $\mathcal{G}:\mathbb{R}^{H/2 \times W/2 \times C} \to \mathbb{R}^{N \times C}$ to downsample $\mathcal{M}_h({\boldsymbol{I}})$ into $N$ patches. Note that the projection $\mathcal{G}$ is implemented by interpolation and flattening operations, without introducing additional trainable parameters. Then,  we  sample a subsequence of patches in ~\cref{one}, which contains   patches with top-$K$ high-freq responses:
\begin{gather}
\small
   \widetilde{\mathbf{X}}^{I}:=\left\{ \mathcal{F}({\boldsymbol{x}^I_j}) | \boldsymbol{x}^I_j \in \textup{top-}K(\lvert\lvert \mathcal{G}\left(\mathcal{M}_h({\boldsymbol{I}})\right) \rvert\rvert_2)\right\}
\label{con:set}
\end{gather}
where $\boldsymbol{x}^I_j$ denotes the $j$-th patch segmented from the IR modality image.
$\widetilde{\mathbf{X}}^{I}$ is the IR high-freq subsequence and $\lvert\lvert \cdot \rvert\rvert_2$ is the ${\ell}_2$-norm.

\textbf{(\romannumeral2) RGB-IR correlated high-freq patch enhancement.}
To capture RGB patches highly correlated with IR high-freq patches, an Exponential Moving Average ViT (EMA-ViT)  $\boldsymbol{G}\left(\theta^{\prime}; \cdot\right)$ is introduced. Specifically, at each training iteration, we use   $\boldsymbol{G}\left(\theta^{\prime} ; \cdot\right)$
to encode both IR  high-freq subsequence $\widetilde{\mathbf{X}}^{I}$ and RGB  entire sequence $\mathbf{X}^{R}$:
\begin{equation}
\small
\widetilde{\mathbf{X}}^{I}=\boldsymbol{G}\left(\theta^{\prime} ; \widetilde{\mathbf{X}}^{I}\right), \mathbf{X}^{R}=\boldsymbol{G}\left(\theta^{\prime} ; \mathbf{X}^{R}\right)
\label{G}
\end{equation}
where the parameters $\theta^{\prime}$ are updated from the Part-based ViT parameters $\theta$: $\theta^{\prime}=m \theta^{\prime}+(1-m) \theta$, and
$m$ denotes the momentum factor close to 1, e.g., 0.9999. The parameters $\theta$ are updated by standard learning optimization.
Given the encoded  $\widetilde{\mathbf{X}}^{I}$ and $\mathbf{X}^{R}$ in ~\cref{G}, we calculate the patch-level  similarity matrix:
\begin{equation}
\small
\boldsymbol{S}_{I\textup{-}R}=\widetilde{\mathbf{X}}^{I} \mathbf{X}^{{R}^{\top}}, \boldsymbol{S}_{I\textup{-}R} \in \mathbb{R}^{M \times N}
\end{equation}
where $M$ and $N$ are the numbers of sampled IR high-freq patches and RGB whole  patches, respectively. Each column vector of $\boldsymbol{S}_{I\textup{-}R}$ indicates how similar each RGB  embedding is to the IR  high-freq patches. Thus, we run the row average of matrix $\boldsymbol{S}_{I\textup{-}R}$ and  generate a similarity vector $\boldsymbol{s} \in \mathbb{R}^{N}$, in which the class token is excluded.
Intuitively, a higher value in $\boldsymbol{s}$ implies that the corresponding RGB  embedding is more correlated with IR high-freq patches.
Hence, we directly select the RGB  patches, which are with top-$K$ values in $\boldsymbol{s}$, as correlated visual patterns with $\widetilde{\mathbf{X}}^{I}$. Consequently, the selected RGB patches constitute the subsequence $\widetilde{\mathbf{X}}^{R}$:
\begin{gather}
\small
   \widetilde{\mathbf{X}}^{R}:=\left\{ \mathcal{F}({\boldsymbol{x}^R_j}) | \boldsymbol{s}_j \in \textup{top-}K(\boldsymbol{s})\right\}
\label{con:set_2}
\end{gather}
where $\boldsymbol{s}_j$ is the $j$-th element of vector $\boldsymbol{s}$.
Afterward, the class token $\boldsymbol{x}_{[\mathrm{CLS}]}$ in \cref{one} is prepended to
the IR subsequence $\widetilde{\mathbf{X}}^{I}$ in ~\cref{con:set} and the RGB subsequence $\widetilde{\mathbf{X}}^{R}$ in ~\cref{con:set_2}, respectively. Then we empower the vanilla ViT to take $\widetilde{\mathbf{X}}^{I}$ and $\widetilde{\mathbf{X}}^{R}$ as  auxiliary input, and the encoded class tokens are served as the IR and RGB high-freq enhanced representations $\boldsymbol{z}^{I}$ and $\boldsymbol{z}^{R}$, respectively.
Note that vanilla ViT and Part-based ViT are weight-sharing during training.
Subsequently,  we optimize the representations $\boldsymbol{z}^{I}$ and $\boldsymbol{z}^{R}$ by minimizing the following objective:
\begin{equation}
\small
\setlength{\abovedisplayskip}{4pt}
\setlength{\belowdisplayskip}{4pt}
\begin{split}
\mathcal{L}_{high}=\mathcal{L}_{CE}\left(\boldsymbol{z}^{I}\right)+\mathcal{L}_{Tri}\left(\boldsymbol{z}^{I}\right)+\mathcal{L}_{CE}\left(\boldsymbol{z}^{R}\right)+\mathcal{L}_{Tri}\left(\boldsymbol{z}^{R}\right)
\label{con:high-freq}
\end{split}
\end{equation}
$\mathcal{L}_{high}$ benefits the ViT to  enhance the representation ability
and identity discrimination
of RGB-IR correlated high-freq components, which are more robust to the distribution shift than holistic images.

\begin{figure*}[h]
  \centering
  \includegraphics[width=0.9\textwidth]{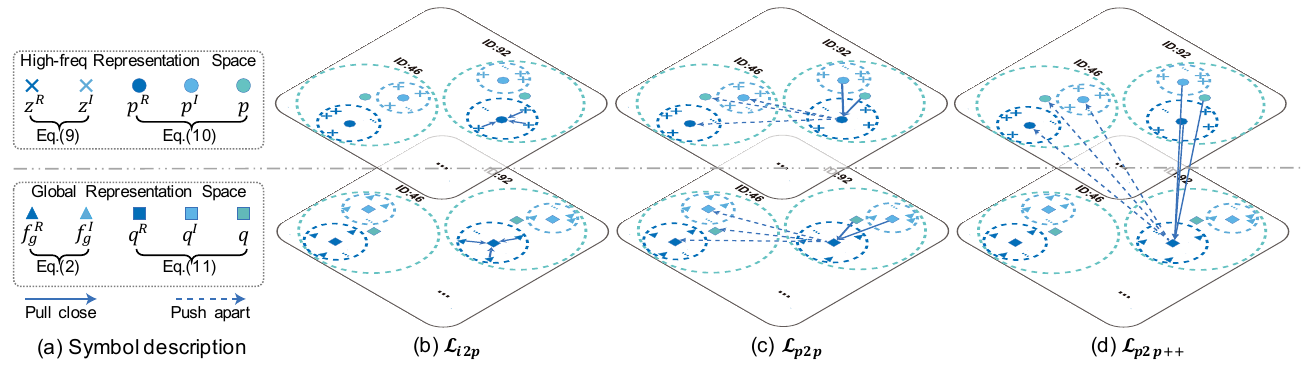}
  \caption{Illustration of MultiProCo. For notational simplicity, we only visualize the supervision on  representations  $\boldsymbol{q}^R$ and $\boldsymbol{p}^R$ of ``ID:92''.
   $\mathcal{L}_{i2p}$ synergizes with $\mathcal{L}_{p2p}$ to hierarchically capture  comprehensive semantics of different modal instances, thus facilitating the aggregation of representations belonging to  the same identity.
  $\mathcal{L}_{p2p++}$ benefits ViT to  capture RGB-IR correlated high-freq components  during inference without  relying on ProtoHPE, thus bringing no extra complexity.}
\label{multiproco}
\end{figure*}

\subsection{Multimodal Prototypical Contrast}
\label{multiproco-section}

In ~\cref{con:high-freq}, $\mathcal{L}_{high}$ optimizes   representations $\boldsymbol{z}^R$ and $\boldsymbol{z}^I$ based only on instances in the mini-batch, ignoring  comprehensive semantics of  instances in different mini-batches. This may cause
inconsistent representations of the same identity during network optimization.

To remedy such inadequacy, we propose Multimodal Prototypical Contrast (MultiProCo), as shown in ~\cref{multiproco}.  Note that a modality prototype is defined as a representative embedding for instances belonging to  the same identity and the same modality. Below, we explain the underlying mechanism of MultiProCo, which consists of \textbf{(\romannumeral1)Multimodal Prototypes Construction and Update}, \textbf{(\romannumeral2)Instance-Prototype Constraint}, and \textbf{(\romannumeral3)Multimodal Prototypes Contrastive Regularization}.

\textbf{(\romannumeral1)Multimodal Prototypes Construction and Update.} To acquire stable multimodal prototypes, we directly construct and dynamically update the prototypes based on statistic centers of representations in each mini-batch. Given the
enhanced RGB-IR correlated
 high-freq
representations $\boldsymbol{z}^I$ and $\boldsymbol{z}^R$, the multimodal high-freq prototypes of the $c$-th identity are formulated below:
\begin{gather}
\small
    \boldsymbol{p}^I_c=\frac{1}{P} \sum_{y_i=c}\boldsymbol{z}_{i}^I,
    \boldsymbol{p}^R_c=\frac{1}{P} \sum_{y_i=c}\boldsymbol{z}_{i}^R,
    \boldsymbol{p}_c=\frac{1}{2P} \sum_{y_i=c}\left(\boldsymbol{z}^I_{i}+\boldsymbol{z}^R_{i}\right)
\label{prototype}
\end{gather}
\begin{gather}
\small
    \boldsymbol{q}^I_c=\frac{1}{P} \sum_{y_i=c}\boldsymbol{f}_{g,i}^I,
    \boldsymbol{q}^R_c=\frac{1}{P} \sum_{y_i=c}\boldsymbol{f}_{g,i}^R,
    \boldsymbol{q}_c=\frac{1}{2P} \sum_{y_i=c}\left(\boldsymbol{f}^I_{g,i}+\boldsymbol{f}^R_{g,i}\right)
\label{prototype}
\end{gather}
where $P$ is the number of IR/RGB  instances per identity  in a mini-batch, $y_i$ is the identity label of the $i$-th instance, ${f}_{g,i}^I$ and ${f}_{g,i}^R$ denote the global representations of the $i$-th IR and RGB instances, respectively.
Then, the above prototypes are dynamically updated by the exponential moving average. Taking the
IR high-freq prototype as an example, we have:
\begin{equation}
\small
\left.\boldsymbol{p}_c^I\right|_{\textup{iter}}:=\left.\alpha \boldsymbol{p}_c^I\right|_{\textup{iter}}+\left.(1-\alpha) \boldsymbol{p}_c^I\right|_{\textup{iter}-1}
\label{prototype-update}
\end{equation}
where $\alpha$ is the exponential decay rate and
``$\textup{iter}$" denotes the current training iteration.

% Multimodal prototypes  benefit the ViT to
% consider comprehensive distributions of different modality instances with the same identity and mitigate the randomness of each mini-batch sampling, thereby stabilizing the training process.

\textbf{(\romannumeral2)Instance-Prototype Constraint.} The purpose is to reduce  intra-modal  variations of instances belonging to the same identity. Specifically, we minimize the distance between  feature representations and their
 corresponding modality prototypes. As shown in ~\cref{multiproco} (b),   the instance-prototype constraint is formulated below:
\begin{equation}
\small
\mathcal{D}\left(\boldsymbol{a},\boldsymbol{b}\right)=\left\|\boldsymbol{a}-\boldsymbol{b}\right\|_2
\end{equation}
\begin{equation}
\small
    \mathcal{L}_{i2p}^h=\frac{1}{P} \sum_{i=1}^P\mathcal{D}\left({z}_i^I,\boldsymbol{p}_{y_i}^I\right)+\frac{1}{P} \sum_{i=1}^P\mathcal{D}\left({z}_i^R,\boldsymbol{p}_{y_i}^R\right)+
    \frac{1}{2P}\sum_{i=1}^{2P}\mathcal{D}\left({z}_i,\boldsymbol{p}_{y_i}\right)
\end{equation}
\begin{equation}
\small
\mathcal{L}_{i2p}^g=\frac{1}{P} \sum_{i=1}^P\mathcal{D}\left({f}_{g,i}^I,\boldsymbol{q}_{y_i}^I\right)+\frac{1}{P} \sum_{i=1}^P\mathcal{D}\left({f}_{g,i}^R,\boldsymbol{q}_{y_i}^R\right)+
    \frac{1}{2P}\sum_{i=1}^{2P}\mathcal{D}\left({f}_{g,i},\boldsymbol{q}_{y_i}\right)
\end{equation}
\begin{equation}
\small
\mathcal{L}_{i2p}=\mathcal{L}_{i2p}^g+\mathcal{L}_{i2p}^h
\label{i2p}
\end{equation}
With $\mathcal{L}_{i2p}$, instances within a modality belonging to the same identity tend to aggregate together. However, $\mathcal{L}_{i2p}$ neglects  diverse semantics and structural  distributions across  different modalities and different identities. To this end, we further propose Multimodal Prototypes Contrastive Regularization below.

\textbf{(\romannumeral3)Multimodal Prototypes Contrastive Regularization.} The objective is to create a representation space where multimodal prototypes belonging to the same identity are pulled together, while prototypes belonging to different identities are pushed apart.

In ~\cref{multiproco} (c), let $\boldsymbol{P}_c=\left[\boldsymbol{p}_c^I, \boldsymbol{p}_c^R, \boldsymbol{p}_c\right]$ and
$\boldsymbol{Q}_c=\left[\boldsymbol{q}_c^I, \boldsymbol{q}_c^R, \boldsymbol{q}_c\right]$
denote the  multimodal high-freq and global prototype sets  of the $c$-th identity, respectively, we have:
% \begin{equation}
% \begin{aligned}
%   S(c, i, k, j)=\exp \left({\frac{\boldsymbol{P}_{c, i}}{\left\|\boldsymbol{P}_{c, i}\right\|_2}}^{\top} \cdot \frac{\boldsymbol{P}_{k, j}}{\left\|\boldsymbol{P}_{k, j}\right\|_2} \bigg/ \tau\right)
% \end{aligned}
% \end{equation}
\begin{equation}
\small
\begin{aligned}
  \mathcal{S}(\boldsymbol{a},\boldsymbol{b})=\exp \left({\frac{\boldsymbol{a}}{\left\|\boldsymbol{a}\right\|_2}}^{\top} \cdot \frac{\boldsymbol{b}}{\left\|\boldsymbol{b}\right\|_2} \bigg/ \tau\right)
\end{aligned}
\label{sim}
\end{equation}
\begin{equation}
\small
\begin{aligned}
\mathcal{L}_{p2p}^h=\frac{1}{C} \sum_{c=1}^C \sum_{i=1} \log \frac{\sum\limits_{j \neq i} \mathcal{S}(\boldsymbol{P}_{c,i},\boldsymbol{P}_{c,j})}{\sum\limits_{j \neq i} \mathcal{S}(\boldsymbol{P}_{c,i},\boldsymbol{P}_{c,j})+\sum\limits_{k \neq c}^C \sum\limits_{j=1} \mathcal{S}(\boldsymbol{P}_{c,i},\boldsymbol{P}_{k,j})}
\label{only_high}
\end{aligned}
\end{equation}
\begin{equation}
\small
\begin{aligned}
\mathcal{L}_{p2p}^g=\frac{1}{C} \sum_{c=1}^C \sum_{i=1} \log \frac{\sum\limits_{j \neq i} \mathcal{S}(\boldsymbol{Q}_{c,i},\boldsymbol{Q}_{c,j})}{\sum\limits_{j \neq i} \mathcal{S}(\boldsymbol{Q}_{c,i},\boldsymbol{Q}_{c,j})+\sum\limits_{k \neq c}^C \sum\limits_{j=1} \mathcal{S}(\boldsymbol{Q}_{c,i},\boldsymbol{Q}_{k,j})}
\label{only_high}
\end{aligned}
\end{equation}
\begin{equation}
\small
\mathcal{L}_{p2p}=\mathcal{L}_{p2p}^g+\mathcal{L}_{p2p}^h
\label{p2p}
\end{equation}
where $\boldsymbol{P}_{c,i}$ and  $\boldsymbol{Q}_{c,i}$ denote the $i$-th element of $\boldsymbol{P}_c$ and $\boldsymbol{Q}_c$, respectively. $\tau$ is the temperature factor and $C$ is the number of identities in the entire training set.

One distinct strength of MultiProCo over traditional instance contrast is that $\mathcal{L}_{i2p}$ synergizes with $\mathcal{L}_{p2p}$ to  hierarchically capture comprehensive semantics of different modal instances. This facilitates the aggregation of global and high-freq enhanced representations belonging to the same identity. With it,  ViT can  learn  compact and informative representations to bridge the gap  robustly.

To prevent RGB-IR correlated high-freq components from being suppressed by  low-freq ones when taking the entire sequence ~\cref{one} as input, we  extend  ~\cref{p2p}  to global and high-freq hybrid contrast:
\begin{equation}
\small
\begin{aligned}
\mathcal{L}_{p2p++}=\frac{1}{C} \sum_{c=1}^C \sum_{i=1} \log \frac{\sum\limits_{j \neq i} \mathcal{S}(\boldsymbol{Q}_{c,i},\textup{sg}(\boldsymbol{P}_{c,j}))}{\sum\limits_{j \neq i} \mathcal{S}(\boldsymbol{Q}_{c,i},\textup{sg}(\boldsymbol{P}_{c,j}))+\sum\limits_{k \neq c}^C \sum\limits_{j=1} \mathcal{S}(\boldsymbol{Q}_{c,i},\textup{sg}(\boldsymbol{P}_{k,j}))}
\label{global_high}
\end{aligned}
\end{equation}
where ``$\textup{sg}$'' denotes the stop-gradient operation.
$\mathcal{L}_{p2p++}$ benefits  ViT to capture RGB-IR correlated high-freq components when taking  the entire sequence ~\cref{one} as input during inference. Hence, ProtoHPE is  necessary during training and can be removed during inference, bringing no extra complexity.
Finally, we optimize ProtoHPE by minimizing the overall objective:
\begin{equation}
\small
\mathcal{L}_{overall}=\mathcal{L}_{base}+\mathcal{L}_{high}+\mathcal{L}_{i2p}+\mathcal{L}_{p2p}+\mathcal{L}_{p2p++}
\end{equation}

\begin{table*}[]
\caption{Comparison with the state-of-the-arts on SYSU-MM01 dataset. Rank-k accuracy and mAP are reported. Optimal and suboptimal results are highlighted in bold and underlined, respectively.}
\label{sysu}
\centering
\resizebox{0.6\textwidth}{!}{
\begin{tabular}{ll|acca|acca}
\toprule
\multicolumn{2}{c|}{Settings}       & \multicolumn{4}{c|}{{\it All-Search}} & \multicolumn{4}{c}{{\it Indoor-Search}} \\ \midrule
\multicolumn{1}{l|}{Method} & Venue & R1 (\%)    & R10 (\%)    & R20 (\%)    & mAP (\%)   & R1 (\%)     & R10 (\%)    & R20 (\%)    & mAP (\%)    \\ \midrule
\multicolumn{1}{l|}{cmGAN~\cite{dai2018cross}}   & IJCAI18      &  26.97     &  67.51      &  80.56  &27.80    &   31.63    &  77.23      &  89.18      &    42.19           \\
\multicolumn{1}{l|}{AliGAN~\cite{wang2019rgb}}       & ICCV19       &  42.40     &  85.00      &   93.70     &   40.7    &  45.90      &  87.60      &  94.40      &  54.30      \\
\multicolumn{1}{l|}{CMSP~\cite{wu2020rgb}}       & IJCV20      &   43.56    &   86.25     &  -      &  44.98     &   48.62     &   89.5    & - &   57.5            \\
\multicolumn{1}{l|}{MACE~\cite{ye2020cross}}       & TIP20      &  51.64     &   87.25     &  94.44      &   50.11    &  57.35      &  93.02      &  97.47      &  64.79      \\
\multicolumn{1}{l|}{HCT~\cite{liu2020parameter}}       & TMM20      &  61.68     &   93.1     &  97.17      &   57.51    &  63.41      &  91.69      &  95.28      & 68.17      \\
\multicolumn{1}{l|}{AGW~\cite{ye2021deep}}       &  TPAMI21     &   47.5    &  84.39      &   92.14     &  47.65     &  54.17      &    91.14    & 95.98       &  62.97      \\
\multicolumn{1}{l|}{MCLNet~\cite{hao2021cross}}       & ICCV21      &  65.40     &  93.3      &    97.14    &   61.98    &   \underline{72.56}     &  96.88      &  99.20      &  \underline{76.58}      \\
\multicolumn{1}{l|}{SMCL~\cite{wei2021syncretic}}       &  ICCV21     &  67.39     &    92.87    &   96.76     &  61.78     &  68.84      &  96.55      & 98.77       &  75.56      \\
\multicolumn{1}{l|}{CM-NAS~\cite{fu2021cm}}     &CVPR21   & 61.99      &  92.87     &   97.25     &   60.02     &   67.01    &  \underline{97.02}      &   \underline{99.32}     &    72.95          \\
\multicolumn{1}{l|}{MID~\cite{zhang2022fmcnet}}       &  AAAI22     &   60.27    &  -      &    -    &  59.40     & 64.86       &    -    & -       &   70.12     \\
\multicolumn{1}{l|}{SPOT~\cite{chen2022structure}}       & TIP22      &  65.34     &   92.73     &  97.04      &  62.25     &  69.42      &  96.22      &  99.12      & 74.63       \\
\multicolumn{1}{l|}{FMCNet~\cite{zhang2022fmcnet}}       &  CVPR22     &   66.74    &  -      &    -    &  62.51     & 68.15       &    -    & -       &   74.09     \\
\multicolumn{1}{l|}{PMT~\cite{lu2022learning}}       & AAAI23    &   \underline{67.53}    &   \underline{95.36}     & \textbf{98.64}       &   \underline{64.98}    &   71.66     &  96.73      &  99.25      &    76.52    \\\midrule

\multicolumn{1}{l|}{\textbf{ProtoHPE}}    & -  &  \textbf{71.92}      &  \textbf{96.19}     &   \underline{97.98}     &  \textbf{70.59}      &   \textbf{77.81}    &  \textbf{98.64}      & \textbf{99.59}       &   \textbf{81.31}           \\ \bottomrule
\end{tabular}}
\end{table*}

\section{Experiment}
\subsection{Dataset and Evaluation Protocol}
\textbf{SYSU-MM01}~\cite{wu2017rgb} is the first large-scale benchmark dataset for VI-ReID. Specifically, it contains 491 pedestrians with total 287,628 visible images and 15,792 infrared images, which are collected by 4 visible and 2 infrared cameras. Four cameras are deployed in the outdoor environments and two are deployed in the indoor environments. The training set contains 395 persons, including 22258 visible images and 11909 infrared images. The test set contains 96 persons, with 3,803 IR images for query and 301/3010 (one-shot/multi-shot) randomly selected RGB images as the gallery. Meanwhile, it contains two different tesing settings, {\it all-search} and {\it indoor-search} settings, in which the {\it all-search} setting uses all images for testing and the {\it indoor-search} mode only uses the indoor images.

\textbf{RegDB}~\cite{wu2017rgb} is collected by a dual-camera system, including one visible and one infrared camera. It contains  412 identities, and each identity has 10 visible and 10 infrared images. Following  previous evaluation protocol~\cite{wei2021syncretic, wu2021discover}, we randomly select all images of 206 identities for training and the remaining 206 identities for testing. The testing stage also contains two evaluation modes, {\it Visible-to-infrared} and {\it Infrared-to-visible} mode. The former means that the model retrieves the person in the infrared gallery when given a visible image, and vice versa. To obtain stable results, we randomly divide this dataset ten times for independent training and testing.

Following conventions~\cite{yan2021beyond,he2020fastreid,he2021transreid}, we adopt
Cumulative Matching Characteristic (CMC) curves and the mean Average Precision (mAP) to evaluate the quality of different methods.
\subsection{Implementation Details}
All the experiments are performed on a single NVIDIA V100 GPU using the PyTorch framework. We adopt  ViT-B/16~\cite{dosovitskiy2020image} as our backbone with initial weights pre-trained on ImageNet. We set the overlap stride to 12 to balance speed and performance and resize all person images to $256\times128$.
The training images are augmented with horizontal flipping and random erasing.
For infrared images, color jitter and gaussian blur are additionally applied. The batch size is set to 64 with a total of 8 different identities. For each identity, 4 visible images and 4 infrared images are sampled. We adopt AdamW optimizer with a cosine annealing learning rate scheduler for training. The basic learning rate is initialized as $3e^{-4}$ and weight decay is set to $1e^{-4}$. The parameter $K$ in ~\cref{con:set} is set to 30\%. The exponential decay rate in ~\cref{prototype-update} and the temperature factor  in ~\cref{sim} are set to 0.8 and 0.1, respectively. The number of parts in the Part-based ViT baseline is set to 4.

\subsection{Comparison with  State-of-the-art Methods}

\textbf{Comparisons on SYSU-MM01.}
We compare ProtoHPE with the state-of-the-art approaches under both all-search and indoor-search settings in ~\cref{sysu}. Specifically, ProtoHPE achieves competitive performance, reaching 71.92\%/70.59\% and 77.81\%/81.31\% Rank-1/mAP in the all-search and indoor-search settings, respectively. From ~\cref{sysu}, we draw the following observations. (1) Compared with PMT~\cite{lu2022learning} which is also based on the ViT framework, ProtoHPE outperforms it in both settings. The root reason is that PMT focuses on learning modality-shared features based on the correlation of RGB and IR holistic person images. This strategy may perform poorly under distribution shifts caused by large differences in wavelength, pose, and background clutter. In contrast,  ProtoHPE effectively bridges the modality gap based on RGB-IR correlated high-freq components. Such components contain discriminative visual patterns and are more robust to the distribution shift than holistic images. (2) Compared to SMCL~\cite{wei2021syncretic}, which reduces the modality gap based only on instances in the mini-batch,
one distinct strength of ProtoHPE  is the ability to capture comprehensive semantics of different modalities. This facilitates the aggregation of representations belonging to  the same identity, thereby learning  compact and informative  representations to robustly bridge the modality  gap.

\textbf{Comparisons on RegDB.}
In ~\cref{RegDB}, ProtoHPE performs favorably against the state-of-the-art methods.
Compared with  DSCNet~\cite{zhang2022dual},  ProtoHPE outperforms it by 3.35\%/6.42\% Rank-1/mAP under the {\it Visible to Infrared} setting. For the more challenging {\it Infrared to Visible} setting, our ProtoHPE also achieves consistent improvement. Instead of  aligning holistic person images, ProtoHPE
bridges the modality gap based on
 RGB-IR correlated high-freq components, e.g.,  heads and human silhouettes, which are more robust to the distribution shift. The above results indicate that   ProtoHPE performs robustly against different datasets and various query settings.

\begin{table}[]
\caption{Comparison with the state-of-the-arts on RegDB.}
\label{RegDB}
\centering
\resizebox{0.9\linewidth}{!}{
\begin{tabular}{ll|cc|cc}
\toprule
\multicolumn{2}{c|}{Settings}       & \multicolumn{2}{c|}{{\it Visible to Infrared}} & \multicolumn{2}{c}{{\it Infrared to Visible}} \\ \midrule
\multicolumn{1}{l|}{Method} & Venue & R1 (\%)  & mAP (\%)   & R1 (\%)     & mAP (\%)    \\ \midrule
\multicolumn{1}{l|}{DG-VAE~\cite{pu2020dual}}       &  ACMMM20     &   72.97     &  71.78     &-         &   -     \\
\multicolumn{1}{l|}{HAT~\cite{ye2020visible}}       &  TIFS20     &   71.83     &  67.56     &70.02         &   66.30     \\
\multicolumn{1}{l|}{MACE~\cite{ye2020cross}}       &  TIP20     &   72.37     &  69.09     &72.12        &   68.57     \\
\multicolumn{1}{l|}{AGW~\cite{ye2021deep}}       &  TPAMI21     &   70.05     &  66.37     &70.49        &   65.90     \\
\multicolumn{1}{l|}{MCLNet~\cite{hao2021cross}}       &  ICCV21     &   80.31     &  73.07     &75.93       &   69.49     \\
\multicolumn{1}{l|}{CM-NAS~\cite{fu2021cm}}       &  ICCV21     &   84.54     &  80.32     &82.57       &   78.31     \\
\multicolumn{1}{l|}{CA~\cite{ye2021channel}}       &  ICCV21     &   85.03     &  79,14     &  \underline{84.75}     &   77.82     \\

\multicolumn{1}{l|}{NFS~\cite{chen2021neural}}       &  CVPR21     &   80.54    &  72.10     &77.95       &   69.97     \\
\multicolumn{1}{l|}{MPANet~\cite{wu2021discover}}       &  CVPR21     &   83.70    &  80.90     &82.80       &   \underline{80.70}     \\
\multicolumn{1}{l|}{SPOT~\cite{chen2022structure}}       &  TIP22     &   80.35    &  72.46     &79.37       &   72.26    \\
\multicolumn{1}{l|}{MSCLNet~\cite{zhang2022modality}}       &  ECCV22     &   84.17    &  \underline{80.99}     &83.86       &   78.31     \\
\multicolumn{1}{l|}{PMT~\cite{lu2022learning}}       &  AAAI23     &   84.83    &  76.55     & 84.16       &   75.13     \\

\multicolumn{1}{l|}{DSCNet~\cite{zhang2022dual}}       &  TIFS23     &   \underline{85.39}    &  77.30     &83.50       &   75.19     \\

\midrule

\multicolumn{1}{l|}{\textbf{ProtoHPE}}        &   -   &      \textbf{88.74}    &  \textbf{83.72}      & \textbf{88.69}    & \textbf{81.99}  \\ \bottomrule
\end{tabular}}
\end{table}

\begin{table}[]
\caption{Ablation study over SYSU-MM01 dataset.}
\label{abla}
\begin{center}
\resizebox{0.9\linewidth}{!}{
\begin{tabular}{c|c|c|ccc|cc}
\toprule
\multirow{2}{*}{Index} & \multirow{2}{*}{Base} & \multirow{2}{*}{CHPE} &  \multicolumn{3}{c|}{MultiProCo} & \multicolumn{2}{c}{SYSU-MM01} \\ \cline{4-8}
                       &                       &                        & $\mathcal{L}_{i2p}$        & $\mathcal{L}_{p2p}$       & $\mathcal{L}_{p2p++}$       & R1 (\%)            & mAP (\%)          \\ \midrule
1                      &  \checkmark                     &                      &           &          &          &  63.11             &   60.34            \\
2                      &  \checkmark                                          &  \checkmark  &       &          &          &       67.50        &      66.79         \\
3                      &  \checkmark                     & \checkmark                      &    \checkmark       &          &          &     68.58          &    68.05           \\
4                      &     \checkmark                  & \checkmark                      &    \checkmark       &     \checkmark    &          &   69.73            &  68.92             \\

5                     &     \checkmark                  & \checkmark                      &    \checkmark       &     \checkmark    &     \checkmark     &      \textbf{71.92}         &      \textbf{70.59}         \\
\bottomrule
\end{tabular}}
\end{center}
\end{table}

\begin{table}[]
\small
\caption{Comparison in ``MultiProCo {\it vs}
 Instance contrast" on SYSU-MM01.}
\label{table:ablation2}
\begin{center}
% \vspace{-1em}
% \setlength{\abovecaptionskip}{0.2cm}
\resizebox{0.9\linewidth}{!}{
\begin{tabular}{c|c|c|c|cc}
\toprule
 \multirow{2}{*}{Index}&\multirow{2}{*}{CHPE} & \multirow{2}{*}{Instance Contrast} &\multirow{2}{*}{MultiProCo} & \multicolumn{2}{c}{SYSU-MM01}   \\
 & & & &  R1 (\%) & mAP (\%)
\\
\midrule
   1&       \checkmark         & \checkmark   & & 69.18 &  66.94\\
 2& \checkmark  &  &\checkmark  & \textbf{71.92}&   \textbf{70.59}  \\
\bottomrule
\end{tabular}}
\end{center}
\end{table}

\subsection{Ablation Study}
We conduct ablation studies on SYSU-MM01 to analyze each core design, including Cross-modal High-freq Patch Enhancement (CHPE) and Multimodal Prototypical Contrast (MultiProCo). We consider the Part-based ViT as Baseline, and results are shown in ~\cref{abla}.

\textbf{Effectiveness of CHPE.}
From index-1 and index-2, CHPE greatly improves the Rank-1/mAP by 4.39\%/6.45\%. This indicates that enhancing RGB-IR correlated high-freq components is  of great significance for boosting VI-ReID.

\textbf{Effectiveness of $\mathcal{L}_{i2p}$ in MultiProCo (\cref{i2p}).}
From index-2 and index-3, $\mathcal{L}_{i2p}$  improves the Rank-1/mAP by 1.08\%/1.26\%. The improvement shows that $\mathcal{L}_{i2p}$ benefits  ViT to reduce the intra-modal variation of  instances belonging to the same identity, thus extracting compact representations.

\textbf{Effectiveness of $\mathcal{L}_{p2p}$ in MultiProCo (\cref{p2p}).}
From index-3 and index-4,  $\mathcal{L}_{p2p}$ elevates the Rank-1/mAP by 1.15\%/0.87\%. This indicates that $\mathcal{L}_{p2p}$ synergizes with $\mathcal{L}_{i2p}$  to
hierarchically capture
comprehensive semantics of different modal instances.
This facilitates aggregating  representations belonging to the same identity, thus robustly
 bridging the modality gap.

\textbf{Effectiveness of $\mathcal{L}_{p2p++}$ in MultiProCo (\cref{global_high}).}
From index-4 and index-5,  $\mathcal{L}_{p2p++}$ improves  the Rank-1/mAP  by  2.19\%/1.67\%. The improvement indicates that  $\mathcal{L}_{p2p++}$ benefits the ViT to capture  RGB-IR correlated high-freq components when taking  the entire sequence ~\cref{one} as input during inference. With it, ProtoHPE is necessary during training and can be removed during inference, without bringing extra complexity.
\begin{figure*}[h]
  \centering
\includegraphics[width=0.72\textwidth]{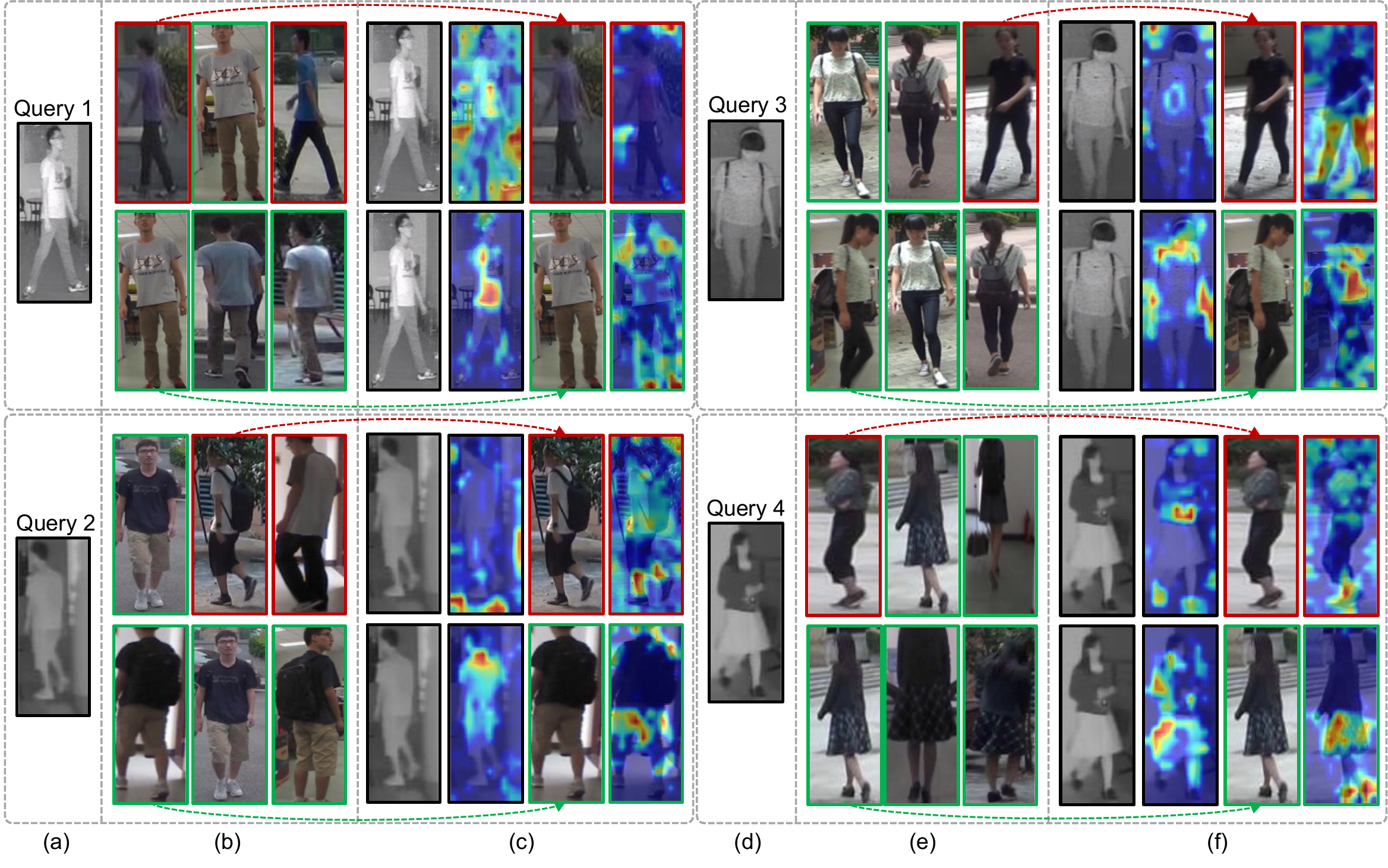}
  \caption{Comparisons in top-3 retrieval results and attention maps between  Baseline and ProtoHPE on SYSU-MM01. Given each IR query in (a) and (d),  the 1st and 2nd rows in (b) and (e) show the top-3 retrieval results of  Baseline and  ProtoHPE, respectively. There are errors in  the retrieval results of  Baseline (highlighted in red boxes), while the results of ProtoPHE are all correct (highlighted in green boxes). In (c) and (f), we visualize the attention maps of  Baseline on the query and wrong retrieval result (the 1st row),  and the attention maps of ProtoHPE on the query and correct retrieval result (the 2nd row).}
\label{top3}
\end{figure*}

\begin{figure}[h]
  \centering
\includegraphics[width=0.85\linewidth]{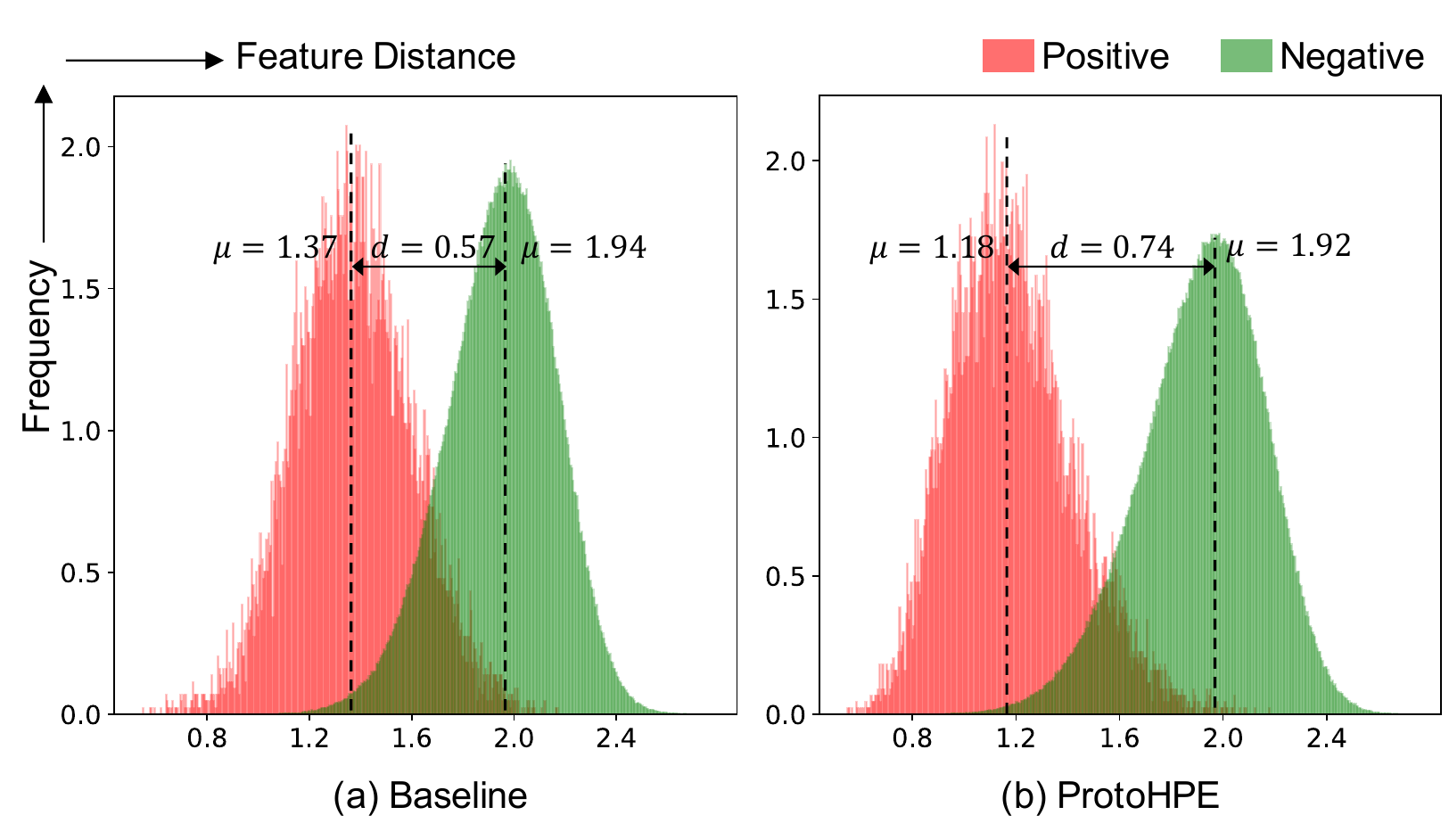}
  \caption{ Euclidean distance distribution of positive and negative  pairs on SYSU-MM01 \textbf{during inference}.}
\label{hist}
\end{figure}
To further verify the effectiveness of  MultiProCo, we  conduct a comparison on ``MultiProCo {\it vs.} Instance  contrast".
We implement it by replacing  MultiProCo with traditional instance contrast, formulated as follows:
\begin{equation}
\small
\begin{aligned}
\mathcal{L}_{inst}^h=\frac{1}{B}  \sum_{i=1}^B \log \frac{\sum\limits_{j:y_j=y_i} \mathcal{S}(\boldsymbol{z}^R_{i},\boldsymbol{z}^I_{j})}{\sum\limits_{j:y_j=y_i} \mathcal{S}(\boldsymbol{z}^R_{i},\boldsymbol{z}^I_{j})+  \sum\limits_{j:y_j \neq y_i}  \mathcal{S}(\boldsymbol{z}^R_{i},\boldsymbol{z}^I_{j})}
\label{instance_contrast_1}
\end{aligned}
\end{equation}
\begin{equation}
\small
\begin{aligned}
\mathcal{L}_{inst}^g=\frac{1}{B}  \sum_{i=1}^B \log \frac{\sum\limits_{j:y_j=y_i} \mathcal{S}(\boldsymbol{f}^R_{g,i},\boldsymbol{f}^I_{g,j})}{\sum\limits_{j:y_j=y_i} \mathcal{S}(\boldsymbol{f}^R_{g,i},\boldsymbol{f}^I_{g,j})+  \sum\limits_{j:y_j \neq y_i}  \mathcal{S}(\boldsymbol{f}^R_{g,i},\boldsymbol{f}^I_{g,j})}
\label{instance_contrast_2}
\end{aligned}
\end{equation}
\begin{equation}
\small
\begin{aligned}
\mathcal{L}_{inst}=\mathcal{L}_{inst}^g+\mathcal{L}_{inst}^h
\label{instance_contrast_4}
\end{aligned}
\end{equation}
where $B$ is the number of IR/RGB  instances  in a mini-batch.
The result in  ~\cref{table:ablation2} shows that compared to $\mathcal{L}_{inst}$ that bridges the modality gap based only on instances in the mini-batch, MultiProCo facilitates extracting more compact and more discriminative representations to bridge the modality gap robustly.

\textbf{Analysis on the distribution of  positive and negative pairs.}\\
~\cref{hist} shows the Euclidean distance distribution of positive and negative  pairs on SYSU-MM01 during inference. Compared to  Baseline,  ProtoHPE reduces the mean distance of positive  pairs from 1.37 to 1.18, while increasing the mean distance between positive and negative distributions  from 0.57 to 0.74. The above result shows that  ProtoHPE benefits  Baseline to effectively reduce the modality gap of instances belonging to the same identity, while pushing instances belonging to different identities away from each other.

\subsection{Visualization Analysis}
\cref{top3} shows comparisons in top-3 retrieval results and attention maps between  Baseline and  ProtoHPE on SYSU-MM01 during inference.
Taking the IR  query ``Query 1'' in ~\cref{top3} (a) as an example, the 1st and 2nd rows in ~\cref{top3} (b) show the top-3 retrieval results of  Baseline and ProtoHPE, respectively. Compared to  Baseline which incorrectly  matches person images with different identities from the query (highlighted in red boxes),  ProtoHPE can correctly identify  person images with the same identity as the query (highlighted in green boxes). To further explain the effectiveness of  ProtoHPE against  Baseline, we utilize Grad-CAM to visualize the attention maps of  Baseline on ``Query 1'' and the wrong retrieval result
(the 1st row in ~\cref{top3} (c)), and  the attention maps of  ProtoPHE on ``Query 1'' and the correct retrieval result (the 2nd row in ~\cref{top3} (c)). We can see that  Baseline mistakenly takes the background clutter as discriminative parts, thereby leading to  wrong retrieval results. In contrast, our ProtoHPE yields lower responses on the background and low-freq components, while focusing on RGB-IR correlated high-freq components, such as heads and human silhouettes.

\section{Conclusion}
In this work, we developed Prototype-guided High-frequency Patch Enhancement, termed ProtoHPE, for visible-infrared person re-identification. We propose  Cross-modal High-freq Patch Enhancement, which  enhances the representation ability of  RGB-IR correlated high-freq components. Such components contain discriminative visual patterns that are less affected by the distribution shift than holistic images. To extract semantically compact and informative representations of the same identity, we  propose Multimodal Prototypical Contrast (MultiProCo). MultiProCo hierarchically captures  comprehensive semantics of different modal instances, facilitating the aggregation of representations belonging to the same identity. Extensive experimental results perform favorably against  mainstream methods on SYSU-MM01 and RegDB datasets.
\textbf{Acknowledgements.} This work was partially supported by the National Natural Science Foundation of China (No. 62072022) and the Fundamental Research Funds for the Central Universities.

%% the bibliography file

\bibliographystyle{ACM-Reference-Format}
\balance
\bibliography{sample-base}

\appendix

\section{Appendix}

\subsection{Effectiveness of CHPE}

\begin{table}[H]
\small
\caption{Comparison in  ``mining RGB patches correlated to IR high-freq patches" versus ``mining RGB patches with top-K high-freq responses" in CHPE.}
\label{table:appendix}
\begin{center}
% \vspace{-1em}
% \setlength{\abovecaptionskip}{0.2cm}
\resizebox{\linewidth}{!}{
\begin{tabular}{l|cc}
\toprule
 \multirow{2}{*}{Method}  & \multicolumn{2}{c}{SYSU-MM01}   \\
  &   R1 (\%) & mAP (\%)
\\
\midrule
         Mining RGB patches correlated with IR high-freq patches & \textbf{67.50} &  \textbf{66.79}\\
  Mining RGB patches with top-k high-freq responses  & 67.03&   65.38  \\
\bottomrule
\end{tabular}}
\end{center}
\end{table}

To further verify the effectiveness of ``CHPE", we compare "mining RGB patches correlated to IR high-freq patches in CHPE" versus "directly mining RGB patches with top-K high-freq responses". In ~\cref{table:appendix}, the Rank-1/mAP drops by 0.47\%/1.41\%. The main reason for the performance drop is that some RGB high-freq components are lost in the IR modality due to large differences in wavelength and scattering between RGB and IR modalities. Enhancing these RGB patches introduces interference, which degrades performance. In contrast, although some RGB high-freq components are lost in the IR modality, there are always RGB patches highly correlated with IR high-freq patches. By computing patch-level similarity matrix with an EMA ViT, we can mine RGB patches correlated with sampled IR high-freq patches. These patches are less affected by variations such as wavelength and scattering than holistic images. Thus, enhancing the representation ability of these correlated high-freq patches benefits to effectively bridge the modality gap.

\subsection{Effectivenss of Stop-gradient Operation}

\begin{table}[H]
\small
\caption{Comparison in ``w/. {\it vs.} w/o stop-gradient operation in Eq. (18) and Eq. (19).}
\label{table:appendix}
\begin{center}
% \vspace{-1em}
% \setlength{\abovecaptionskip}{0.2cm}
\resizebox{\linewidth}{!}{
\begin{tabular}{l|cc}
\toprule
 \multirow{2}{*}{Method}  & \multicolumn{2}{c}{SYSU-MM01}   \\
  &   R1 (\%) & mAP (\%)
\\
\midrule
         w/o stop-gradient in Eq. (18) and Eq. (19) & \textbf{71.92} &  \textbf{70.59}\\
  w/ stop-gradient in Eq. (18) and Eq. (19)  & 70.73&  68.27  \\
\bottomrule
\end{tabular}}
\end{center}
\end{table}

The objective of the stop-gradient operation in Eq. (21) is to prevent cross-modal correlated high-frequency components from being disturbed by low-frequency ones in the global representation
 . As shown in Figure 1 of the manuscript, visually similar IR low-frequency patches can represent different semantics, causing feature representations to lose identity discrimination. The stop-gradient operation benefits the global representation $\boldsymbol{f}_g$ to capture discriminative and robust high-frequency components while mitigating low-frequency distractions. In contrast, the purpose of Eq. (18) and Eq. (19) is to capture comprehensive semantics of different modality instances. If a stop-gradient operation is applied, the propagation of gradient items between different modality instances will be reduced, which is not conducive to stable interactions between structural distributions of different modalities.
To further verify it, we add a comparison between "with vs without stop-gradient in Eq. (18) and Eq. (19)". From the result shown in the table below, we can see that the Rank-1/mAP drops by 1.19\%/1.32\%, which validates our analysis.

\subsection{Visualization}

\begin{figure}[H]
  \centering
\includegraphics[width=0.9\linewidth]{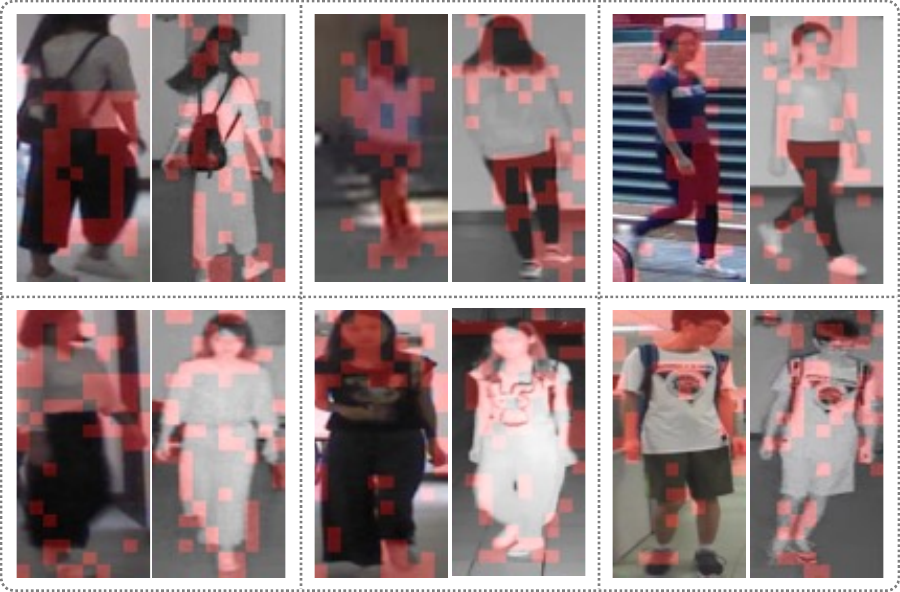}
  \caption{ Visualization of RGB-IR correlated high-frequency patches mined by CHPE.}
\label{chpe}
\end{figure}

\cref{chpe} shows RGB-IR correlated high-frequency patches mined by Cross-modal High-frequency Patch Enhancement (CHPE). We can see that the selected patches mainly reflect
pivotal high-frequency components of person images, such as  heads, cloth textures, and human silhouettes. These patches are less affected by variations such as wavelength, pose, and background clutter than holistic person images, and thus are more
  robust to  the modality gap.

\subsection{Overall Training Procedure}

\cref{alg:algorithm1} shows the overall training procedure of ProtoHPE. Let $\mathcal{V}_{train}$ and $\mathcal{R}_{train}$ denote the training datasets of visible and infrared images, respectively.
At each episodic training iteration, the Part-based ViT Baseline takes the entire sequence  ( Eq. (1) of the manuscript ) as input to enhance the discriminative power of  both global and part representations.
Subsequently, we utilize Cross-modal High-freq Patch Enhancement to enhance the representation ability of RGB-IR correlated high-freq patches. To obtain semantically compact and discriminative high-freq representations, we jointly utilize $\mathcal{L}_{i2p}$ and $\mathcal{L}_{p2p}$, which facilitates the aggregation of representations belonging to the same identity. Furthermore, we use $\mathcal{L}_{p2p++}$ to prevent RGB-IR correlated high-freq components from being suppressed by low-freq ones when taking the entire sequence Eq. (1) as input during network optimization. By virtue of $\mathcal{L}_{p2p++}$, ProtoHPE is only necessary during training and can be removed during inference.
Finally, we optimize ProtoHPE by minimizing the overall objective function in Eq. (22). During inference, only the Part-based ViT Baseline is necessary and our ProtoHPE can be removed, thus bringing no additional complexity.

\begin{algorithm}[t]
	\caption{The overall training procedure of ProtoHPE.}
	\label{alg:algorithm1}
	\KwIn{Training datasets $\mathcal{V}_{train}$ and $\mathcal{R}_{train}$}
	\KwOut{The trained Part-based ViT Baseline.}
	\For{epoch=1 to MaxEpochs}{
		\For{iter=1 to MaxIters}{
			Sample a mini-batch $\mathcal{V}_\mathcal{B} \subset \mathcal{V}_{train}$,   $\mathcal{R}_\mathcal{B} \subset \mathcal{R}_{train}$;\\
            \tcp{Part-based ViT Baseline}
            Enhance  global and part representations by Eq. (2);\\
            \tcp{Cross-modal High-freq Patch Enhancement}
		Mine RGB-IR correlated high-freq patches by Eq. (4)-Eq. (8);

            Enhance the representation ability of RGB-IR correlated high-freq patches by Eq. (9);\\
            \tcp{Multimodal Prototypical Contrast}

            Construct and dynamically update multimodal prototypes by Eq. (10)-Eq. (12);

            Use $\mathcal{L}_{i2p}$ to reduce intra-modal variations of instances belonging to the same identity;

            Use $\mathcal{L}_{p2p}$ to capture comprehensive semantics of different modal instances;

            Use $\mathcal{L}_{p2p++}$ to prevent RGB-IR correlated high-freq patches from being suppressed by low-freq ones;\\

            \tcp{Overall objective function}
            Optimize the network by minimizing the overall objective function in Eq. (22).
		}
	}
\end{algorithm}

\end{document}